\documentclass[runningheads]{llncs}
\usepackage[T1]{fontenc}
\usepackage[table]{xcolor}
\usepackage[dvipsnames]{xcolor}
\usepackage[colorlinks=true, linkcolor=red, citecolor=brown]{hyperref}
\usepackage{graphicx}
\usepackage{orcidlink}
\usepackage{algorithm}
\usepackage{algorithmic}
\usepackage{booktabs}
\usepackage{multirow}
\usepackage{multicol}
\usepackage{caption}
\usepackage{bbding}
\usepackage{makecell}
\usepackage{amsmath}
\usepackage{amssymb}
\usepackage{verbatim}
\usepackage{adjustbox}
\usepackage{fvextra}
\usepackage[utf8]{inputenc}
\usepackage{enumitem}
\usepackage{breakcites}

\usepackage{color}

\begin{document}
\title{Visual Instruction-Finetuned Language Model for Versatile Brain MR Image Tasks}
\newcommand{\fs}{\footnotesize}

\titlerunning{Visual Instruction-Finetuned LM for Brain MRI}
\author{Jonghun Kim\inst{1}\orcidlink{0009-0002-2790-2090} \and
Sinyoung Ra\inst{2} \and
Hyunjin Park\inst{1}\orcidlink{0000-0001-5681-8918}}
\authorrunning{J. Kim et al.}
\institute{Department of Electrical and Computer Engineering, \\
    Sungkyunkwan University, Suwon, Korea \and
    Department of Artificial Intelligence, 
    Sungkyunkwan University, Suwon, Korea \\
\tt\small \{iproj2,nsy0527,hyunjinp\}@skku.edu}
\maketitle              % typeset the header of the contribution

\begin{abstract}
LLMs have demonstrated remarkable capabilities in linguistic reasoning and are increasingly adept at vision-language tasks. The integration of image tokens into transformers has enabled direct visual input and output, advancing research from image-to-text descriptions to text-to-image generation. However, simple text-to-image generation holds limited clinical utility. In medical imaging, tasks such as image segmentation for localizing pathologies or image translation for reconstructing missing sequences have much greater clinical importance. Despite this, integrating these diverse, clinically relevant tasks within a single, versatile language model remains unexplored. Our method, \textbf{LLaBIT} (Large Language Model for Brain Image Translation), extends the visual reasoning of LLMs to these clinically meaningful tasks in the brain MRI domain. To mitigate the spatial information loss inherent in image tokenization, we incorporate a mechanism to reuse feature maps from the image encoder, minimizing data degradation. We also generate text data using LLMs with strict predefined instructions to augment limited image-text paired data in brain MRI. We comprehensively evaluated our method on five brain MRI datasets across four distinct tasks: report generation, visual question answering, image segmentation, and image translation. Our model not only demonstrated superior performance across all tasks but also outperformed specialized, task-specific models in direct comparisons, highlighting its efficacy and versatility. Our code is available at \href{https://github.com/jongdory/LLaBIT}{github.com/jongdory/LLaBIT}.
\end{abstract}
    
\begin{figure} [t]
    \vspace{-6pt}
    \centering
    \includegraphics[width=0.99\columnwidth]{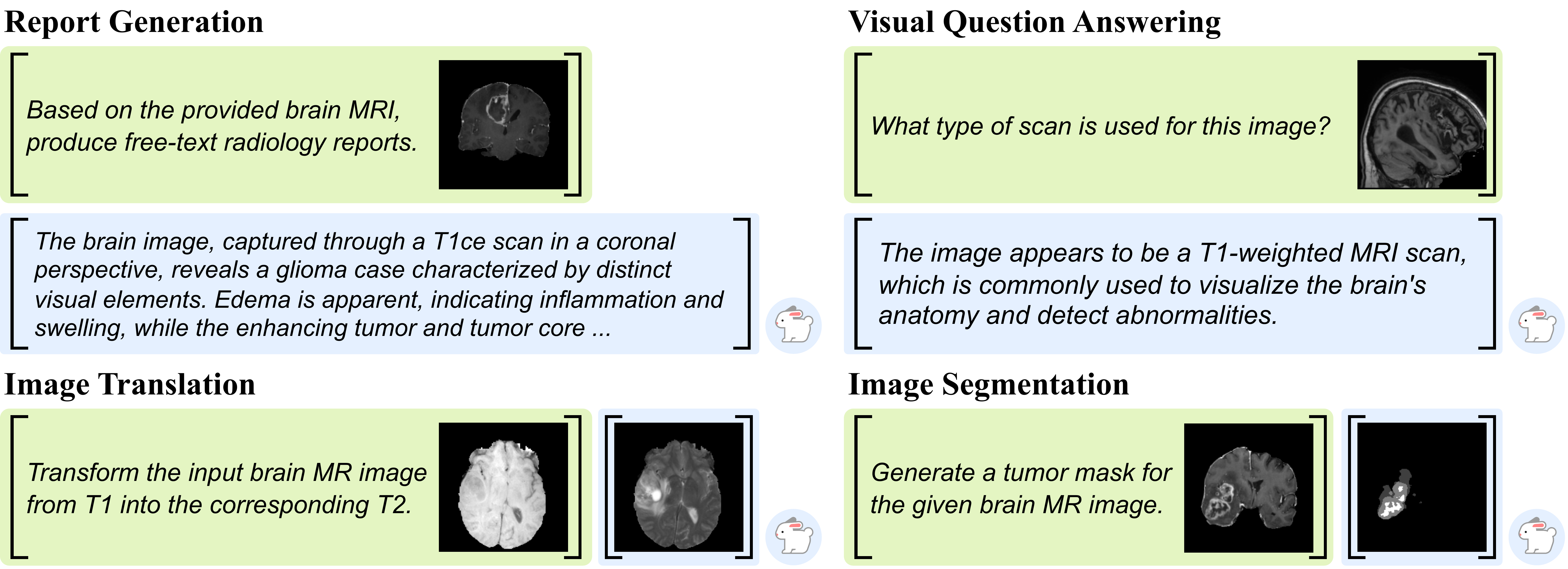}
    \centering
    \vspace{-3pt}
    \caption{\textbf{Example of LLaBIT performing versatile tasks on brain MR images}. LLaBIT supports report generation and image-to-image tasks.}
    \label{fig1}
    \vspace{-6pt}
\end{figure}

\section{Introduction}
\label{sec:intro}

In recent years, large language models (LLMs) have made significant progress \cite{brown2020language,achiam2023gpt}. LLMs have powerful abilities to understand and reason with natural language, allowing them to adapt to different tasks without being specifically trained for each one. As a next step, there have been attempts to incorporate visual information into language models, enabling them to interpret images with language and to ask and answer questions about visual content \cite{achiam2023gpt,liu2023visual,alayrac2022flamingo}. Methods are being developed for language models to not only process visual information as input, but also to produce it as output \cite{koh2023generating}. These methods have great potential in the medical field, where image understanding requires context, reasoning, and the ability to communicate findings in both text and image. A key method for adapting LLMs to new tasks without losing their original abilities is instruction tuning \cite{brown2020language,wei2022finetuned}. This has proven effective for teaching a model to handle diverse instructions and flexibly apply its knowledge. This idea has been extended to visual instruction tuning \cite{liu2023visual}, enabling LLMs to perform conversational discussions about image content. 
Thus, multimodal LLMs that handle both images and language have emerged \cite{liu2023visual,zhu2024minigpt,chaves2024towards}. The main challenge for these models is to align image features with their powerful, pretrained language-derived features without catastrophic forgetting \cite{koh2023generating,zhu2024minigpt}. This task is even harder with medical images than with natural images, because medical images often appear similar. The model must be able to detect subtle but critical differences in order to provide an accurate description. Furthermore, the region of interest, such as a lesion, is often a small, complex part of the whole image, which makes the task even more difficult. Recent work has explored using VQ-GAN \cite{esser2021taming} to transform medical images like chest X-rays (CXRs) into tokens, integrating them into an LLM without changing its structure \cite{lee2024llmcxr}. This allowed the model to generate both text descriptions and their corresponding images. However, simply generating an image that corresponds to a textual description may have limited clinical use and compressing an image into tokens can lead to a loss of visual details.

In brain magnetic resonance imaging (MRI), diagnosing lesions requires precise decisions from multiple perspectives, often achieved by using different imaging modalities or sequences. These sequences are differentiated by physical acquisition parameters like repetition time and echo time. For instance, T1-weighted images excel at delineating normal anatomy, while T2-weighted images are more sensitive for detecting lesions, as abnormalities like water, edema, and tumors appear hyper-intense \cite{damadian1971tumor,bradley1984comparison}. Leveraging information from two or more sequences enables more precise clinical decisions \cite{cha2006update,ellingson2015consensus}. However, in clinical practice, key sequences may be unavailable due to acquisition time constraints or poor image quality. This necessitates medical image translation (MIT), the task of generating a missing sequence from available ones \cite{kim2024adaptive}. MIT is more clinically relevant than generating images from scratch, aided by LLM guidance \cite{lee2024llmcxr}. Furthermore, identifying the precise location and extent of a lesion through segmentation is critical for accurate diagnosis \cite{kwon2024leveraging,kim2025weakly,kim2025enhancing,kwon2025blood}. Segmentation is also imaging task that can benefit from LLM to segment a region based on a textual description. 

To this end, we propose a method that leverages instruction fine-tuning to align medical images and text, enabling an LLM to perform language-driven visual analysis without compromising its inherent linguistic abilities. Our primary motivation is to create an integrated model that goes beyond simple report generation. We aim to empower a language model to describe images or lesions from the perspective of different MR sequences. Our approach provides a single pretrained language model with four key capabilities: report generation, visual question answering, image translation, and image segmentation for brain MRI. To mitigate information loss from image compression, we propose reusing the encoder features from the VQ-GAN for the two image-to-image tasks. We hypothesize that reusing these feature maps preserves critical image details and thereby improves performance on image-to-image tasks. Our results show that our unified model outperforms specialized models designed for each task, demonstrating the significant potential of a text-and-image integrated language model. Representative samples of our model, Large Language Model for Brain Image Translation (\textbf{LLaBIT}), are shown in Figure \ref{fig1}.

\noindent \textbf{Our contributions}
\begin{itemize}[noitemsep, topsep=0pt, partopsep=0pt, parsep=0pt]
    \item[1.] We integrate images into our language model to give it the ability to translate and segment images without losing its language reasoning ability. 
    \item[2.] We reuse the feature map of the encoder along with a zero-skip connection to improve the quality of image-to-image tasks.
    \item[3.] We evaluated our model on text generation, visual question answering, image translation, and image segmentation across five datasets that include brain lesions such as tumors. Our model achieved superior performance in all tasks, showing that a language model can be effectively utilized for not only image-to-text but also image-to-image tasks.
\end{itemize}
\begin{figure} [t]
    \vspace{-6pt}
    \centering
    \includegraphics[width=0.99\textwidth]{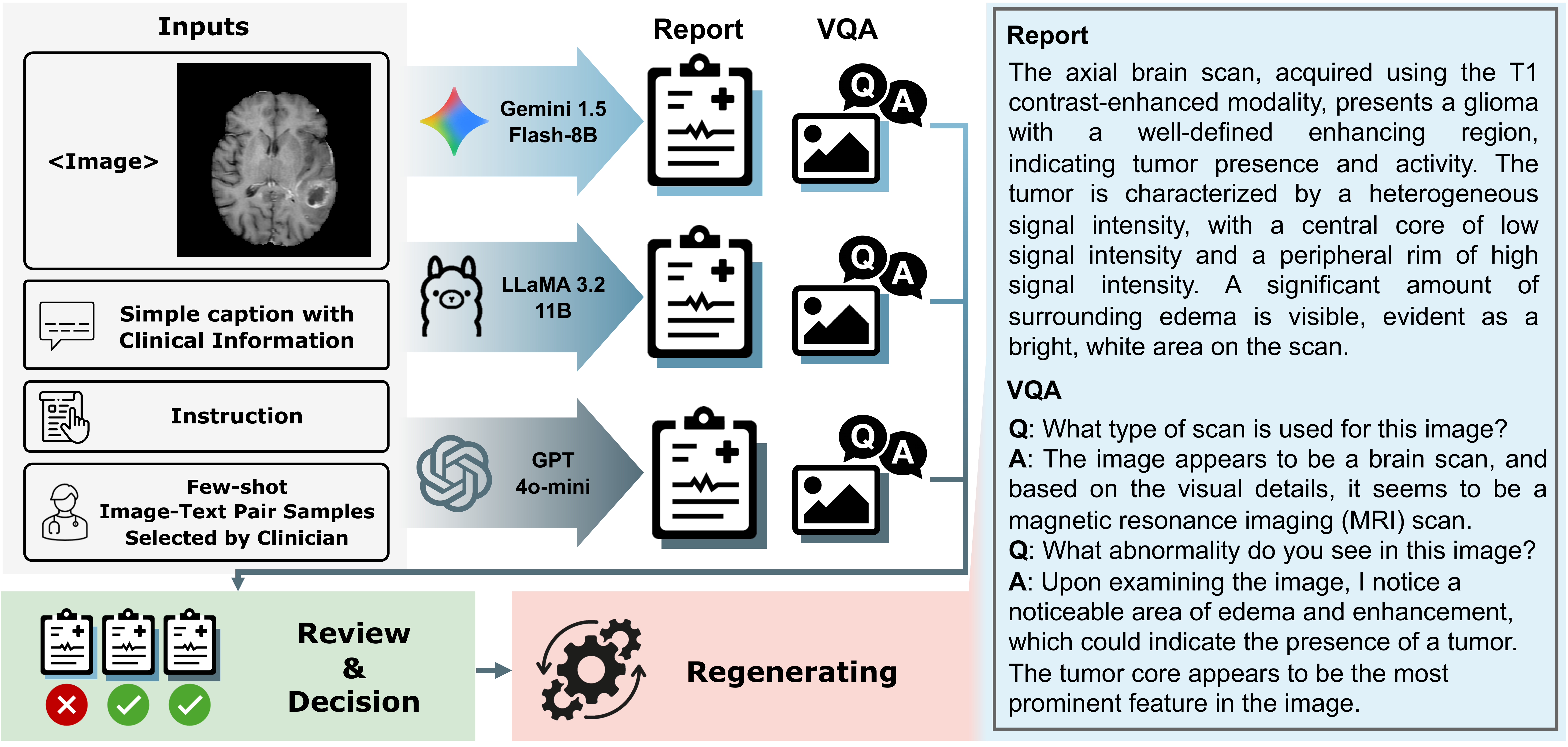}
    \centering
    \vspace{-3pt}
    \caption{\textbf{Text data generation with LLMs on a dataset with only images.} Images and captions are processed by LLMs with strict predefined instructions and few-shot samples selected by clinicians to generate reports and VQA results. The output of each model is accepted or rejected using GPT 4o and the final report and VQA are regenerated based on this feedback.}
    \vspace{-6pt}
    \label{fig2}
\end{figure}

\begin{figure*} [t]
    \vspace{-6pt}
    \centering
    \includegraphics[width=0.99\textwidth]{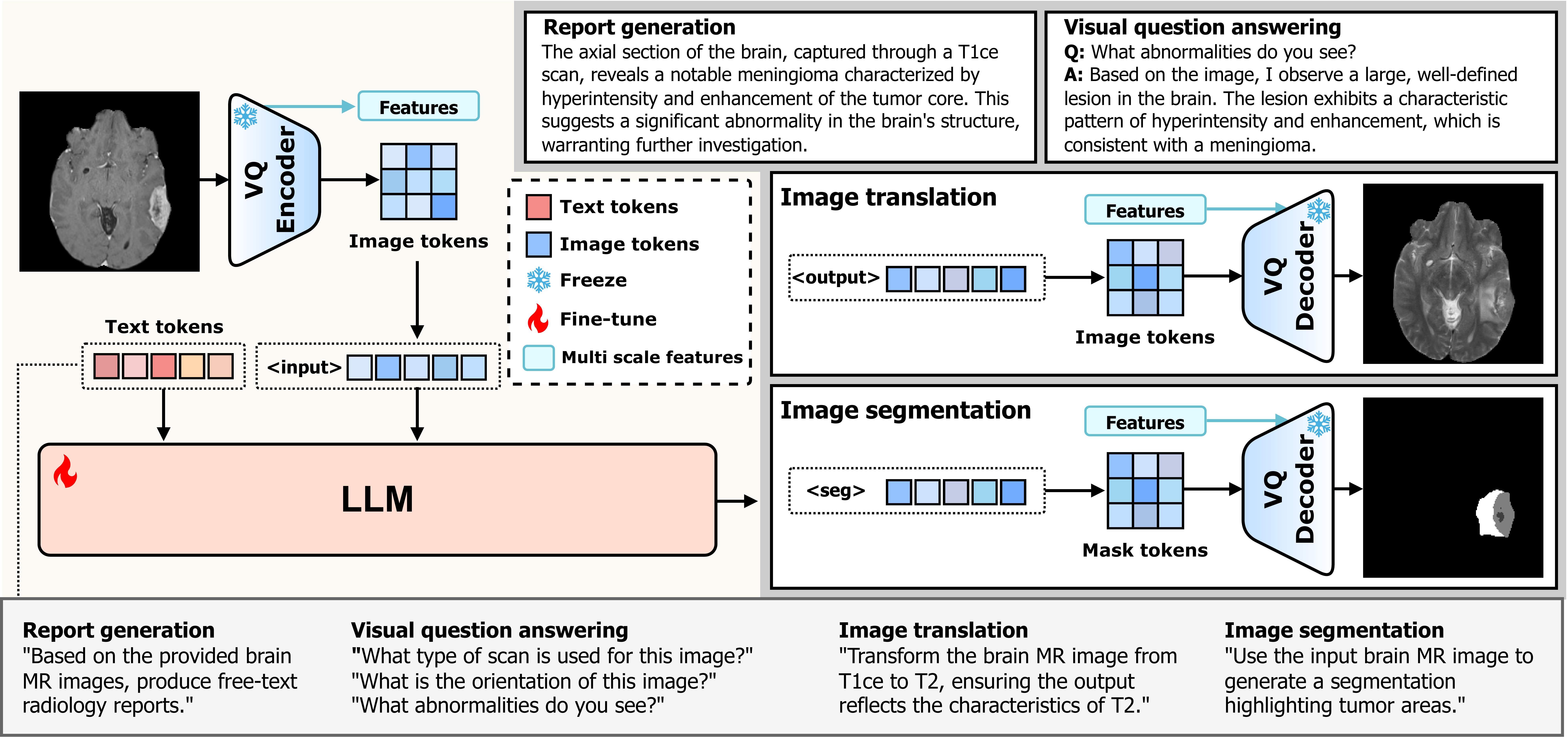}
    \vspace{-3pt}
    \caption{\textbf{Instruction tuning pipeline.} Both text and images are tokenized and fed into the LLM, which can generate either text or image tokens as output. The LLM's vocabulary is extended to include image tokens in addition to text tokens. The instruction is provided to the LLM as text tokens. The image is converted into quantized tokens using a VQ encoder and fed into the LLM along with an \texttt{<input>} token. These image tokens are added to the LLM's vocabulary as discrete values, similar to text tokens. For image translation, the output is generated with an \texttt{<output>} token, while for segmentation, the output is generated with a \texttt{<seg>} token.}
    \vspace{-6pt}
    \label{fig3}
\end{figure*}

\section{Related Work}
\noindent \textbf{Language Models for Medicine.} 
LLMs pretrained on large amounts of text data have made remarkable progress and have been applied to various downstream tasks such as translation, question answering, and text generation \cite{achiam2023gpt,team2023gemini,touvron2023llama,touvron2023llama2openfoundation,grattafiori2024llama3herdmodels,ra2025enhancing}. LLaVA \cite{liu2023visual} successfully integrates image and text information by generating VQA data using image and object location information and applying instruction tuning, even when image-text pairs are limited. In the medical domain, models such as LLaVA-Med \cite{li2024llava} and BioMed-VITAL \cite{cui2024biomedical} have demonstrated the potential to generate captions for medical images and assist in explanation and diagnosis. In addition, medical multimodal LLMs have begun to emerge recently \cite{moor2023med,thawkar2023xraygpt,chaves2024towards,lee2025cxr}. Specifically, LLM-CXR \cite{lee2024llmcxr} proposes a unified model that supports both image-to-text and text-to-image functions for chest X-rays, highlighting its versatility. However, most of these studies have focused on CXR datasets, where paired image-text data is available, while similar research on brain MRI is rare.

\vspace{3pt}
\noindent \textbf{Image-to-Image Tasks with Language Models.}
In image segmentation, several methods combine language models with pretrained CLIP \cite{radford2021learning} or introduce a \texttt{<SEG>} token into LLaVA \cite{liu2023visual}. M3D \cite{bai2024m3d} integrates medical images into large language models such as LLaMA2 \cite{touvron2023llama2openfoundation} to perform both segmentation and text generation tasks.
In image translation, methods that synthesize images by incorporating natural language descriptions using pretrained CLIP have been actively studied. DiffuseIT \cite{kwon2023diffusionbased} integrates text information directly into the noise reduction steps of the diffusion process so that the generated image follows the text prompt. ControlNet \cite{zhang2023adding} leverages Stable Diffusion and adds an extra branch that uses a control image and text prompts to guide image synthesis. However, these approaches are task-specific models designed for image synthesis utilizing embeddings of language models.

\section{Method}
\subsection{Instruction-Driven Text Data Generation}
\label{sec3.1}
One of our main goals is to perform image-to-image tasks based on text instructions without losing the linguistic reasoning capabilities inherent in language models, rather than only generating accurate clinical reports. However, a common problem is that many medical datasets lack image-text pairs, which means we need to generate text data using available clinical information. Nevertheless, generated text data may contain misinformation or hallucinations. To minimize this risk, we build on previous work on generating medical text data with LLMs (shown in Figure \ref{fig2}). We guide the LLMs by providing instructions that include clinician-selected demonstrations from BioMed-VITAL \cite{cui2024biomedical}, along with a few image-text examples selected by clinicians. Recent research has shown that having several models work together can greatly reduce hallucinations \cite{du2023improving}, so we use multiple LLMs, including Gemini 1.5 \cite{team2023gemini}, LLaMA 3.2 \cite{grattafiori2024llama3herdmodels}, and GPT 4o-mini \cite{achiam2023gpt}, to reduce risk. The generated text is then reviewed and scored by GPT-4o. It is subsequently regenerated using this feedback, and the final text data are used as reports in the VQA set. The prompts used for this process and details are described in the Appendix.

\subsection{Stage I: Instruction Tuning}
LLMs are pretrained on large amounts of text data, making it very costly to retrain them from scratch for multimodal tasks. Therefore, we fine-tune a text-based LLM to learn visual information as well. Depending on the instruction, the response corresponding to an image can be generated as text, an image, or a segmentation mask by a single LLM. The image is also tokenized like text and fed to the LLM. Output image tokens are converted back to images by the VQ decoder. The pipeline is shown in Figure \ref{fig3}.

\vspace{3pt}
\noindent \textbf{Image Tokenization.}
To preserve the LLM's language ability during fine-tuning, we add image tokens to its existing vocabulary. For example, if the original LLM has an embedding table of size $\mathbb{R}^{K_\texttt{text}\times d_e}$ , where $d_e$ is the embedding dimension, the table is expanded to $\mathbb{R}^{(K_\texttt{text}+K_\texttt{img})\times d_e}$, where $K_\texttt{text}$ and $K_\texttt{img}$ represent the vocabulary size of text and image tokens, respectively. We use a VQ-GAN to tokenize images by compressing them into quantized latents. The VQ-GAN consists of an encoder $E$ and decoder $D$ as follows:
\begin{equation}
    E(\cdot):\mathbb{R}^{C\times H\times W} \rightarrow \{1,2,\ldots,K_\texttt{img}\}^{d_z}
\end{equation}
\begin{equation}
    D(\cdot):\{1,2,\ldots,K_\texttt{img}\}^{d_z} \rightarrow \mathbb{R}^{C\times H\times W}
\end{equation}
Using VQ-GAN, a tokenized latent $z$ of length $d_z$ is obtained with $z \in \{1,2, \\ \ldots,K_\texttt{img}\}$. The VQ-GAN is trained with the loss defined below:
\begin{equation} 
\mathcal{L}_{VQGAN} = \underbrace{||x - \hat{x}||^2}_{\text{Reconstruction}} + \underbrace{\mathcal{L}_{GAN}}_\text{GAN} + \underbrace{\mathcal{L}_{VQ}}_\text{Codebook}, 
\end{equation}
where $\mathcal{L}_{GAN}$ is the GAN loss \cite{isola2017image} and $\mathcal{L}_{VQ}$ is the quantization loss \cite{van2017neural,esser2021taming}. The VQ encoder and decoder are frozen during LLM training.

\vspace{3pt}
\noindent \textbf{Image to Text Generation.}
We train our model by applying instruction fine-tuning to an LLM pretrained on a large text corpus. This process only fine-tunes the LLM, where no structural changes are made to the model other than expanding the token embedding table. In this task, tokenized images are fed directly into the LLM so that the model learns visual information besides its language abilities. The model is then instructed to generate a report corresponding to the given image. The input image tokens are fed into the LLM along with the \texttt{<input>} token and the model produces a response corresponding to the instruction. The instructions for fine-tuning are provided in the Appendix. An example of text generation is shown below.

\vspace{6pt}
\begin{adjustbox}{scale=0.73}
\begin{minipage}{1.35\linewidth}
\begin{Verbatim}[frame=single, numbersep=5pt,
                 commandchars=\\\{\}]
\textbf{Instruction}: Generate free-text radiology reports for the provided brain MR images.
\textbf{Input}: \texttt{<input>} \texttt{<VQ172 VQ281 ... VQ511 VQ172>}
\textbf{Response}: The axial brain scan, captured using the T1-weighted modality, presents a ...
\end{Verbatim}
\end{minipage}
\end{adjustbox}
\vspace{3pt}

\vspace{3pt}
\noindent \textbf{Image to Image Translation.}
This task aims to generate a target modality or segmentation mask that matches the given instruction by processing image tokens. The input image tokens are provided with an \texttt{<input>} token and the output image tokens are produced with an \texttt{<output>} token. Since the LLM produces image tokens in the same way it generates text tokens, no extra module is required. For the segmentation task, the \texttt{<seg>} token is used instead of the \texttt{<output>} token. An example of image translation task is shown below.

\vspace{6pt}
\begin{adjustbox}{scale=0.73}
\begin{minipage}{1.35\linewidth}
\begin{Verbatim}[frame=single, numbersep=5pt,
                 commandchars=\\\{\}]
\textbf{Instruction}: Generate a brain MR image in T2  based on the T1 input.
\textbf{Input}: \texttt{<input>} \texttt{<VQ172 VQ092 ... VQ658 VQ172>}
\textbf{Response}: \texttt{<output>} \texttt{<VQ172 VQ335 ... VQ079 VQ172>}
\end{Verbatim}
\end{minipage}
\end{adjustbox}
\vspace{3pt}

\vspace{3pt}
\noindent \textbf{Training Objective.} 
The LLM is optimized using an autoregressive objective based on instruction-input pairs \cite{radford2018improving,brown2020language,liu2023visual}. For an image $\textbf{X}_\texttt{img}$ and an instruction $\textbf{X}_\texttt{instruct}$, our objective is to generate a response $\textbf{X}_\texttt{response}$ in an autoregressive manner. The loss is applied only to the tokens generated after the response according to the instruction tuning approach \cite{taori2023stanford,conover2023free}.
For a sequence of length $L$, the training loss is given by:
\begin{equation} 
\begin{split}
\mathcal{L}_\texttt{instruct} = -\log(p(\textbf{X}_\texttt{response}|\textbf{X}_\texttt{img},\textbf{X}_\texttt{instruct})) \\
= \sum_{i=1}^{L}p_\theta(\textbf{x}_i|\textbf{X}_\texttt{img}, \textbf{X}_\texttt{instruct}, \textbf{X}_{\texttt{response}, < i} ),
\end{split}
\end{equation}
where $\theta$ denotes the learnable parameters, and $\textbf{X}_{\texttt{response}, < i}$ denotes the response tokens before the current prediction token $\textbf{x}_i$.

\begin{figure} [t]
    \vspace{-6pt}
    \centering
    \includegraphics[width=0.85\columnwidth]{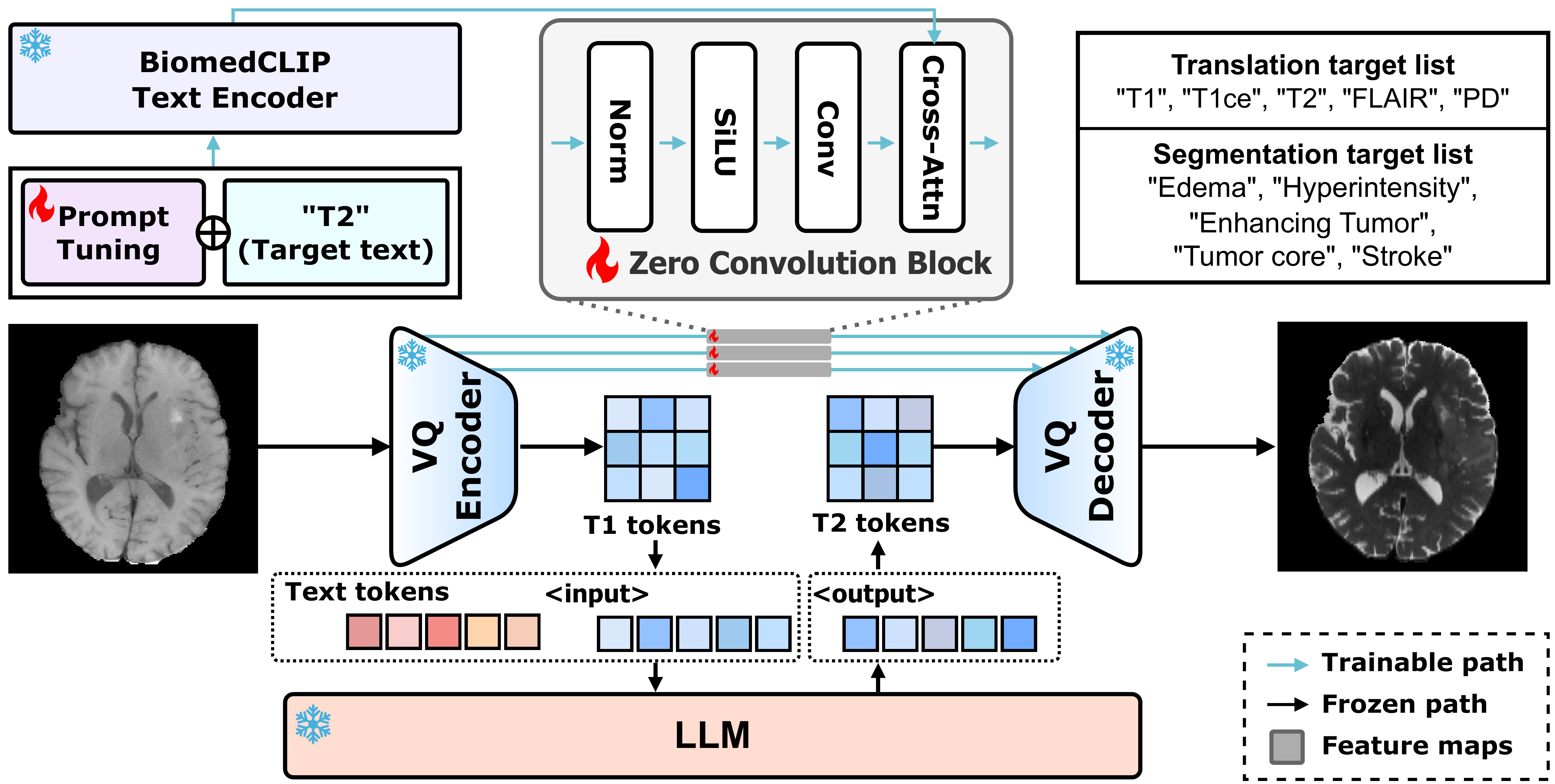}
    \vspace{-3pt}
    \caption{\textbf{Fine-tuning of VQ-GAN with zero skip connection}. The skip connection is fine-tuned, while freezing the image encoder and decoder. A zero convolution block is adopted using a BiomedCLIP text encoder and prompt tuning to flexibly adapt to the target.}
    \vspace{-6pt}
    \label{fig4}
\end{figure}

\subsection{Stage II: Skip Connection Fine-tuning}
The LLM alone can perform image translation with only Stage I instruction tuning, but compressing images into latents makes it difficult to capture fine details \cite{esser2021taming,rombach2022high}. The skip connection in U-Net \cite{ronneberger2015u} preserves fine-scale information with minimal loss that works effectively in image translation. Inspired by this, we add a learnable skip connection for image translation that preserves fine-scale information (shown in Figure \ref{fig4}).

\begin{figure} [t]
    \vspace{-6pt}
    \centering
    \includegraphics[width=0.9\columnwidth]{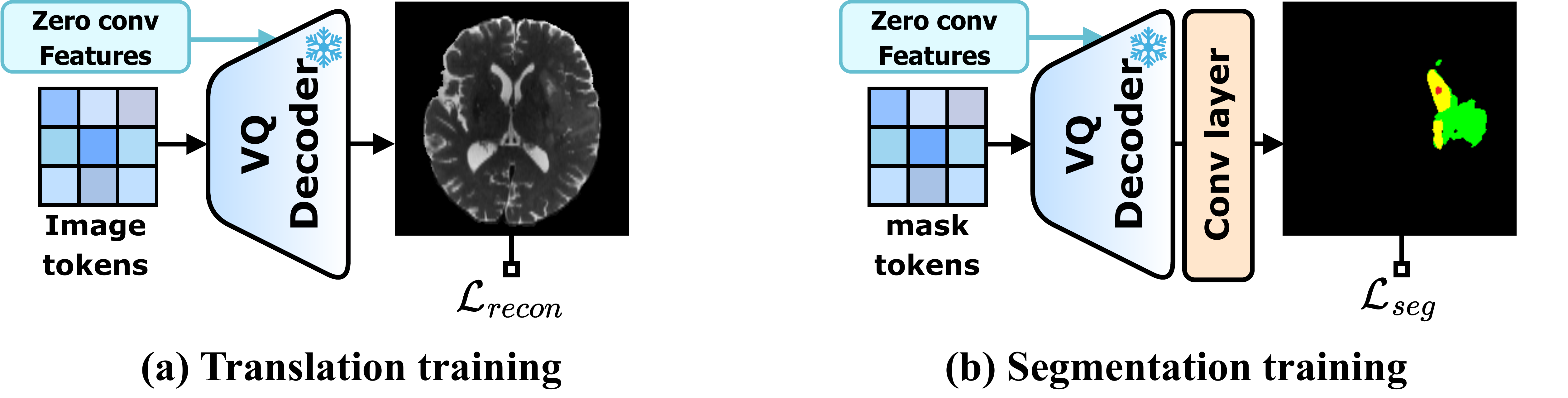}
    \vspace{-3pt}
    \caption{\textbf{Loss functions for image-to-image tasks.} (a) The translation task is trained using reconstruction loss. (b) The segmentation task is trained using Dice loss with an additional layer.}
    \vspace{-6pt}
    \label{fig5}
\end{figure}

\vspace{3pt}
\noindent \textbf{Zero Convolution Block.}
Adding a skip connection initialized with random weights may disrupt the decoding process by causing significant changes in the feature map. We employ zero-initialized convolutions on each feature map, allowing them to be adapted gradually. To flexibly drive the features to the target, we incorporate a BioMedCLIP \cite{zhang2023biomedclip} text encoder along with prompt tuning \cite{jia2022visual,kim2026visual}. We define the prompt $\mathcal{P}$ as follows:
\begin{equation}
    \mathcal{P} = \text{Concat}(P(y_t),\text{CLIP}_\texttt{token}(y_t)),
\end{equation}
where $\text{Concat}$ is the concatenation operation, $P$ represents the learnable prompt, $\text{CLIP}_\texttt{token}$ is the token generated by the BioMedCLIP tokenizer, and $y_t$ is the target text. This prompt is fed into the text encoder and interacts with the image features through cross-attention. We then compute the feature of the  $i$-th skip connection as follows:
\begin{equation}
    f^\texttt{skip}_i = \text{ZeroConv}(f^\texttt{enc}_{i}, \mathcal{P}),
\end{equation}
where $f^\texttt{enc}_{i}$ denotes the $i$-th encoder feature, $\text{ZeroConv}$ represents the zero convolution block, and $f^\texttt{skip}_i$ is the resulting $i$-th skip-connection feature. The decoder feature is updated by adding the output of the $i$-th decoder layer to the corresponding skip-connection feature.

\vspace{3pt}
\noindent \textbf{Loss Function.} Since the VQ decoder produces continuous-valued images, an additional convolution layer is added to the segmentation task to generate class probabilities. We use the reconstruction loss $\mathcal{L}_{recon}$ and the segmentation loss $\mathcal{L}_{seg}$ as follows (shown in Figure \ref{fig5}):
\begin{equation}
    \mathcal{L}_{recon} = ||y-\hat{y}||^2, \ \ \mathcal{L}_{seg} = \text{DICE}(y, \hat{y}),
\end{equation}
where $y$ denotes the target and $\hat{y}$ represents the prediction.

\section{Experiments}

\noindent \textbf{Datasets.} We used the following brain MRI datasets for training and evaluation (\# of \textcolor{RoyalBlue}{train} / \textcolor{ForestGreen}{validation} / \textcolor{Orange}{test} subjects): BraTS2021 \cite{baid2021rsna} (\textcolor{RoyalBlue}{1101} / \textcolor{ForestGreen}{40} / \textcolor{Orange}{110}), BraTS2023-MEN \cite{labella2023asnrmiccai} (\textcolor{RoyalBlue}{880} / \textcolor{ForestGreen}{29} / \textcolor{Orange}{91}), IXI \footnote{https://brain-development.org/ixi-dataset/} (\textcolor{RoyalBlue}{500} / \textcolor{ForestGreen}{27} / \textcolor{Orange}{51}), and ATLAS 2.0 \cite{liew2022large} (\textcolor{RoyalBlue}{430} / \textcolor{ForestGreen}{23} / \textcolor{Orange}{43}). BraTS2021 and BraTS2023-MEN include T1, T1ce, T2, and FLAIR modalities; IXI contains T1, T2, and PD modalities; and ATLAS 2.0 consists of the T1 modality. We also employed UPENN-GBM \cite{bakas2021multi} (\textcolor{Orange}{147}) for an external test dataset (image-to-image tasks only).

\vspace{3pt}
\noindent \textbf{Implementation Details.} We used \texttt{dolly-v2-3b} \cite{conover2023free} as our backbone language model, which has a vocabulary size of 50,821 tokens ($K_\texttt{text}$). For the image tokenizer, we used VQGAN \cite{esser2021taming} with a codebook $K_\texttt{image}$ of size 1024. By adding 1024 image tokens, we extended the embedding table to a total of 51,845 tokens ($K_\texttt{text} + K_\texttt{image}$). We used ROUGE \cite{lin-2004-rouge}, METEOR \cite{banerjee-lavie-2005-meteor}, and BERT-F1 \cite{Zhang2020BERTScore} to evaluate report generation. For segmentation, we measured performance using the Dice score. For image translation, we used PSNR, SSIM, and FID \cite{heusel2017gans} as metrics. The model was trained on four A100 80GB GPUs with a batch size of 2 per GPU, a learning rate of 5e-6, and the AdamW \cite{loshchilov2018decoupled} optimizer. Additional implementation details are provided in the Appendix.

\begin{figure}[t]
\vspace{-6pt}
  \centering
  \includegraphics[width=0.99\textwidth]{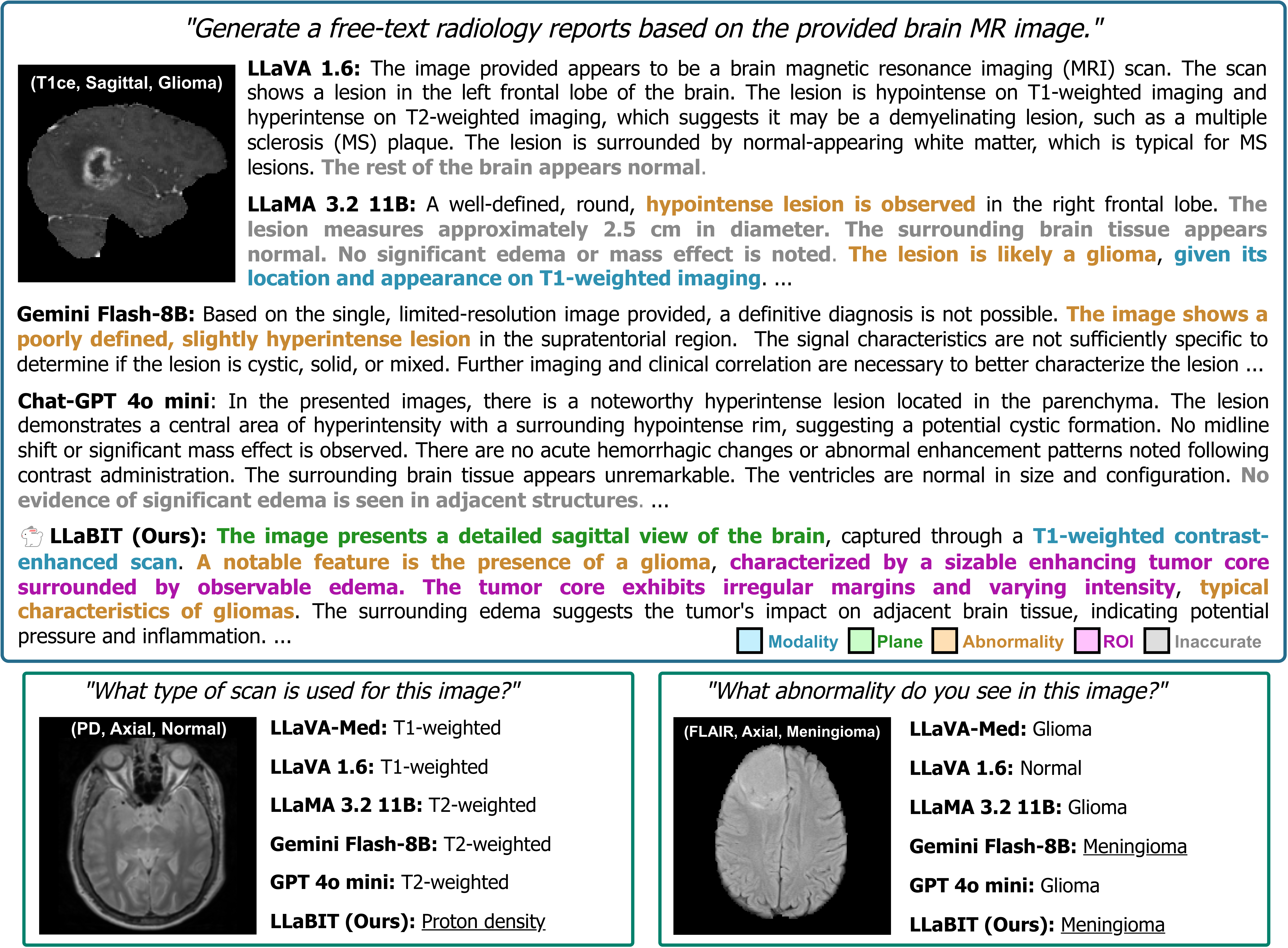}
  \vspace{-3pt}
    \captionof{figure}{\textbf{Examples of report generation and VQA from various MLLMs}. The highlighted text reflects the unique features of each image. For comparison, each image displays its modality, plane, and abnormality information at the top. Top: an example of report generation; Bottom: an example of VQA.}
    \label{fig6}
    \vspace{-6pt}
\end{figure}

\begin{table}[t]
  \vspace{-6pt}
  \centering
  \caption{\textbf{Performance on image-to-text tasks}. Top: Report generation performance evaluated across four datasets. Higher scores on the ROUGE, METEOR, and BERT-F1 metrics indicate better performance. Bottom: VQA performance on four datasets. "Mod." represents modality and "Abno." represents abnormality, with each value representing accuracy. \textbf{Bold} denotes the best performance, and * indicates a significant difference between ours and baselines ($p$-value$\ < 0.05$).}
  \label{table1}
  \vspace{3pt}
  \setlength{\tabcolsep}{2pt} 
  \centering
  \scalebox{0.725}{
    \centering
    \begin{tabular}{l|ccc|ccc|ccc|ccc}
        \toprule
        \multirow{2}{*}{Report Generation} &  \multicolumn{3}{c|}{BraTS2021}  & \multicolumn{3}{c|}{BraTS2023-MEN}  &  \multicolumn{3}{c|}{ATLAS 2.0} & \multicolumn{3}{c}{IXI}  \\
         & \small R. & \small M. & \small B. & \small R. & \small M. & \small B. & \small R. & \small M. & \small B. & \small R. & \small M. & \small B. \\
        \cmidrule(lr){1-13}
        \small LLaVA-Med                                 &  7.19  \ & 12.37 \ & 85.52 \ &  6.26 \ & 11.38 \ & 85.43 \ &  7.15 \ & 14.37 \ & 86.74 \ &  9.25 \ & 19.70 \ & 87.69 \ \\
        \small LLaVA 1.6                                 & 20.66  \ & 19.98 \ & 84.24 \ & 21.48 \ & 21.10 \ & 84.34 \ & 25.05 \ & 26.00 \ & 86.40 \ & 28.56 \ & 30.59 \ & 87.80 \ \\
        \small LLama 3.2 11B                             & 15.78  \ & 15.32 \ & 83.10 \ & 16.43 \ & 16.52 \ & 83.15 \ & 16.62 \ & 15.20 \ & 83.07 \ & 17.12 \ & 16.04 \ & 83.32 \ \\
        \small Gemini 1.5 8B                       & 22.04  \ & 17.69 \ & 81.94 \ & 22.01 \ & 17.28 \ & 81.67 \ & 21.63 \ & 17.66 \ & 81.90 \ & 22.65 \ & 18.30 \ & 82.06 \ \\
        \small GPT 4o-mini                               & 17.42  \ & 15.86 \ & 81.55 \ & 16.13 \ & 14.99 \ & 81.39 \ & 18.97 \ & 18.23 \ & 82.02 \ & 22.28 \ & 20.27 \ & 82.73 \ \\
        \rowcolor{gray!10} \small \textbf{LLaBIT (Ours)} & \textbf{35.03}$^*$ & \textbf{39.35}$^*$ & \textbf{89.71}$^*$ & \textbf{33.36}$^*$ & \textbf{36.84}$^*$ & \textbf{89.12}$^*$ & \textbf{33.69}$^*$ & \textbf{36.97}$^*$ & \textbf{89.44}$^*$ & \textbf{37.33}$^*$ & \textbf{41.82}$^*$ & \textbf{90.50}$^*$ \\
        \bottomrule
        \toprule
        \multirow{2}{*}{VQA} &  \multicolumn{3}{c|}{BraTS2021}  & \multicolumn{3}{c|}{BraTS2023-MEN}  &  \multicolumn{3}{c|}{ATLAS 2.0} & \multicolumn{3}{c}{IXI}  \\
         & Mo. & Pl. & Ab. & Mo. & Pl. & Ab. & Mo. & Pl. & Ab. & Mo. & Pl. & Ab.  \\
        \midrule
        \small LLaVA-Med                                 & 0.736 \ & 0.391 \ & 0.682 \ & 0.747 \ & 0.385 \ & 0.582 \ & 0.907 \ & 0.349 \ & 0.279 \ & 0.902 \ & 0.373 \ & 0.216 \ \\
        \small LLaVA 1.6                                 & 0.291 \ & 0.336 \ & 0.536 \ & 0.286 \ & 0.308 \ & 0.286 \ & 0.605 \ & 0.256 \ & 0.186 \ & 0.431 \ & 0.392 \ & 0.745 \ \\
        \small LLama 3.2 11B                             & 0.864 \ & 0.922 \ & 0.800 \ & 0.846 \ & 0.923 \ & 0.780 \ & 0.907 \ & 0.744 \ & 0.674 \ & 0.863 \ & 0.902 \ & 0.922 \ \\
        \small Gemini 1.5 8B                       & 0.664 \ & 0.745 \ & 0.500 \ & 0.670 \ & 0.681 \ & 0.549 \ & 0.767 \ & 0.791 \ & 0.605 \ & 0.745 \ & 0.804 \ & 0.902 \ \\
        \small GPT 4o-mini                               & 0.482 \ & 0.518 \ & 0.182 \ & 0.527 \ & 0.560 \ & 0.154 \ & 0.744 \ & 0.907 \ & 0.186 \ & 0.549 \ & 0.922 \ & 0.608 \ \\
        \rowcolor{gray!10} \small \textbf{LLaBIT (Ours)} & \textbf{0.945}$^*$ & \textbf{0.927}$^*$ & \textbf{0.891}$^*$ & \textbf{0.945}$^*$ & \textbf{0.945}$^*$ & \textbf{0.934}$^*$ & \textbf{0.977}$^*$ & \textbf{0.953}$^*$ & \textbf{0.721}$^*$ & \textbf{0.922}$^*$ & \textbf{0.961}$^*$ & \textbf{0.980}$^*$ \\
        \bottomrule
    \end{tabular}
    }
    \vspace{-6pt}
\end{table}
  
\begin{table}[t]
    \vspace{-6pt}
    \centering
    \caption{\textbf{Performance on image-to-image tasks}. "WT." denotes whole tumor, "ET." denotes enhancing tumor, "TC." denotes tumor core, and "Avg." represents the average value. Each value is expressed as a dice coefficient. \textbf{Bold} indicates the best performance, \underline{underlined} denotes the second best performance, and * significant difference between ours and baselines, $p$-value$\ < 0.05$.}
    \label{table2}
    \vspace{3pt}
  
  \begin{minipage}{1\linewidth}
    \setlength{\tabcolsep}{1.8pt} 
    \centering
    \scalebox{0.65}{
        \begin{tabular}{l|cccc|cccc|cccc|c}
            \toprule
            \multirow{2}{*}{Image Segmentation} &  \multicolumn{4}{c|}{BraTS2021}  & \multicolumn{4}{c|}{BraTS2023-MEN} & \multicolumn{4}{c|}{UPENN-GBM} & ATLAS 2.0\\
             & WT. & ET. & TC. & Avg. & WT. & ET. & TC. & Avg. &  WT. & ET. & TC. & Avg. & \fs \textbf{Stroke}\\
            \cmidrule(lr){1-14}
            M3D (Pretrained)              & 0.025 \  & 0.024 \  & 0.010 \  & 0.019 \  & 0.013 \  & 0.008 \  & 0.008 \  & 0.010 \  & 0.041 \  & 0.035 \ & 0.016  \ & 0.030  \ & 0.010 \  \\
            M3D (Fine-tuned)              & 0.608 \  & 0.450 \  & 0.342 \  & 0.466 \  & 0.275 \  & 0.231 \  & 0.101 \  & 0.202 \  & 0.619 \  & 0.411 \  & 0.322 \  & 0.450 \  & 0.057 \  \\
            \rowcolor{gray!10} \textbf{LLaBIT (Stage I)}    & 0.653 \  & 0.445 \  & 0.389 \  & 0.495 \  & 0.492 \  & 0.373 \  & 0.235 \  & 0.366 \  & 0.645 \  & 0.439 \  & 0.318 \  & 0.467 \  & 0.269 \  \\
            \rowcolor{gray!10} \textbf{LLaBIT (Stage II)}    & \textbf{0.770}$^*$ & \textbf{0.747}$^*$ & \textbf{0.603}$^*$ & \textbf{0.706}$^*$ & \textbf{0.651}$^*$ & \textbf{0.587}$^*$ & \textbf{0.402}$^*$ & \textbf{0.546}$^*$ & \textbf{0.792}$^*$ & \textbf{0.778}$^*$ & \textbf{0.613}$^*$ & \textbf{0.727}$^*$ & \textbf{0.438}$^*$ \\
            \bottomrule
        \end{tabular}
    } 
    \end{minipage}
    
    \vspace{1pt}
    
    \setlength{\tabcolsep}{3.2pt} 
    \centering
    \scalebox{0.65}{
        \begin{tabular}{l|ccc|ccc|ccc|ccc}
            \toprule
            \multirow{2}{*}{T1 $\rightarrow$ T2} &  \multicolumn{3}{c|}{BraTS2021}  & \multicolumn{3}{c|}{BraTS2023-MEN} & \multicolumn{3}{c|}{IXI} & \multicolumn{3}{c}{UPENN-GBM} \\
             & \fs PSNR $\uparrow$ & \fs SSIM $\uparrow$ & \fs FID $\downarrow$ & \fs PSNR $\uparrow$ & \fs SSIM $\uparrow$ & \fs FID $\downarrow$ & \fs PSNR $\uparrow$ & \fs SSIM $\uparrow$ & \fs FID $\downarrow$ & \fs PSNR $\uparrow$ & \fs SSIM $\uparrow$ & \fs FID $\downarrow$ \\
            \cmidrule(lr){1-13}
            DiffuseIT                                    & \underline{22.61} & \textbf{0.817}    & \underline{38.42} & 23.40 \              & 0.838             & 44.37             & \underline{22.93} \ & \underline{0.746} \ & 30.09 \             & 21.22 \             & \underline{0.783} \ & 49.76 \             \\
            ControlNet                                   & 22.16             & 0.810             & \textbf{32.35}    & \underline{23.57} \  & \underline{0.845} & \textbf{41.82}    & 22.64 \             & 0.712 \             & \underline{26.13} \ & \underline{22.19} \ & 0.781 \             & \underline{44.20} \ \\
            \rowcolor{gray!10} \textbf{LLaBIT (Stage I)} & 21.23             & 0.722             & 47.99             & 22.42 \              & 0.765             & 51.80             & 21.55 \             & 0.648 \             & 27.54 \             & 21.02 \             & 0.707 \             & 45.38 \             \\
            \rowcolor{gray!10} \textbf{LLaBIT (Stage II)} & \textbf{22.99}    & \underline{0.815} & 42.83             & \textbf{24.32}$^*$   & \textbf{0.846}    & \underline{43.56} & \textbf{23.93}$^*$  & \textbf{0.793}$^*$  & \textbf{24.82}$^*$  & \textbf{22.94}$^*$  & \textbf{0.817}$^*$  & \textbf{43.20}$^*$  \\
            \bottomrule
        \end{tabular}
    }
    \centering
    \vspace{-6pt}
\end{table}

\section{Results}
\noindent \textbf{Report Generation.} We compared the report generation performance of our model with that of other vision-language models that accept images as input along with instructions. The results show that our model achieved the best scores on the ROUGE, METEOR, and BERT-F1 metrics (Table \ref{table1} top). Fig. \ref{fig6} left shows representative outputs from each model. While the reports from the other models contain unnecessary or inaccurate information and lack detailed image descriptions, our model captures all the essential information. This suggests that instruction tuning of the language model with our data generation better captures rich image details. Note that this evaluation is inherently limited, as it relies on LLM-generated data instead of clinician-written reports. Nevertheless, the primary objective of this study was not to produce a flawless clinical report generator, but to demonstrate that a single fine-tuned LLM can be proficient in both image-to-image and image-to-text tasks.

\vspace{3pt}
\noindent \textbf{Visual Question Answering.} We evaluated the VQA performance of our model and other models using three question types to measure image understanding. The questions focus on modality, plane, and abnormality. They are posed in a closed-ended format to ensure that the models provide definite answers. An example of these VQA results is shown in Fig. \ref{fig6} right. We measured answer accuracy of correct answers across four datasets and three question types. The results for each model are summarized in Table \ref{table1} bottom. Our model outperformed other models on every metric, demonstrating an enhanced understanding of brain MR images in terms of modality, plane, and abnormality. 

\begin{figure} [t]
    \centering
    \begin{minipage}{0.48\columnwidth}
        \centering
        \includegraphics[width=\columnwidth]{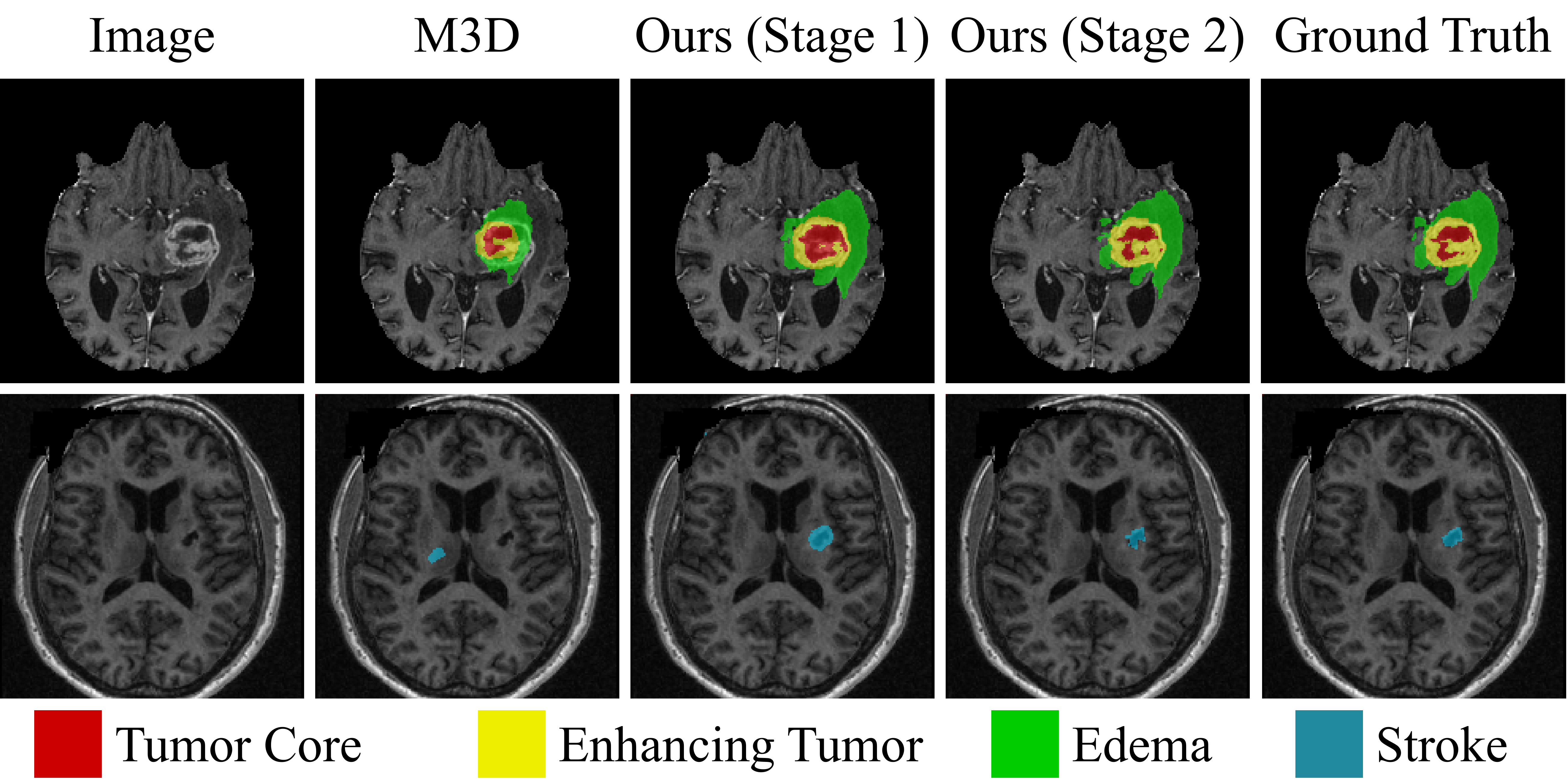}
        \vspace{-6pt}
        \caption{\textbf{Segmentation results}. Top:  BraTS-2021. Bottom: ATLAS 2.0.}
        \label{fig7}
        \end{minipage}
        \begin{minipage} {0.08\textwidth}
        \end{minipage}
        \begin{minipage}{0.48\columnwidth}
        \centering
        \includegraphics[width=\columnwidth]{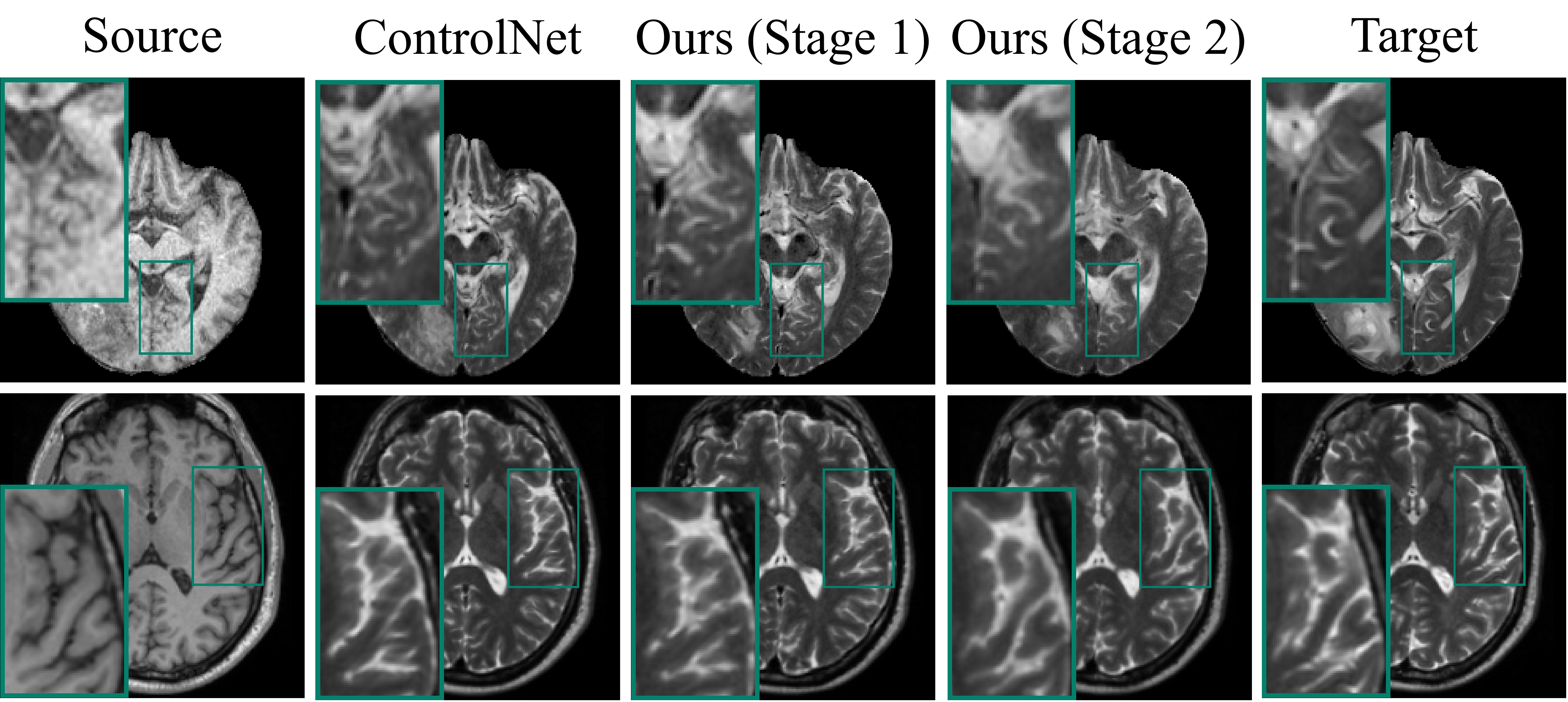}
        \vspace{-6pt}
        \caption{\textbf{T1 $\rightarrow$ T2 translation results.} Top: UPENN-GBM. Bottom: IXI.}
        \label{fig8}
    \end{minipage}
\end{figure}

\vspace{3pt}
\noindent \textbf{Segmentation.} We compared the segmentation performance of our method with M3D, a language model that can perform segmentation (Table \ref{table2} top). Unlike M3D, which relies on an additional segmentation module (such as SegVol \cite{du2024segvol}), our approach uses a single transformer that generates segmentation tokens autoregressively. 
Since the pretrained M3D, which has been trained on various medical datasets, has not seen brain MR images, its performance suffers in brain tumor segmentation. For a fair comparison, we also fine-tuned M3D on the same training dataset as ours. The segmentation results, as shown in Fig. \ref{fig7}, indicate that while the segmentation tokens generated in Stage I are able to roughly segment the lesion area, the Stage II module adds much more detail.

\vspace{3pt}
\noindent \textbf{Image Translation.} Since no previous study has addressed image translation using a language model, we trained and compared models that use text encoders, such as those from CLIP encoders pretrained via contrastive learning on image data, on the same image-text paired dataset. Quantitative results are shown in Table \ref{table2} bottom and qualitative results are shown in Fig. \ref{fig8}. Although it is challenging to outperform models specifically designed for image synthesis with strong inductive bias, our method achieved competitive performance. Furthermore, skip connections led to a noticeable improvement, suggesting that they help to capture fine details more effectively.

\begin{table} [t]
    \centering
    \includegraphics[width=0.65\columnwidth]{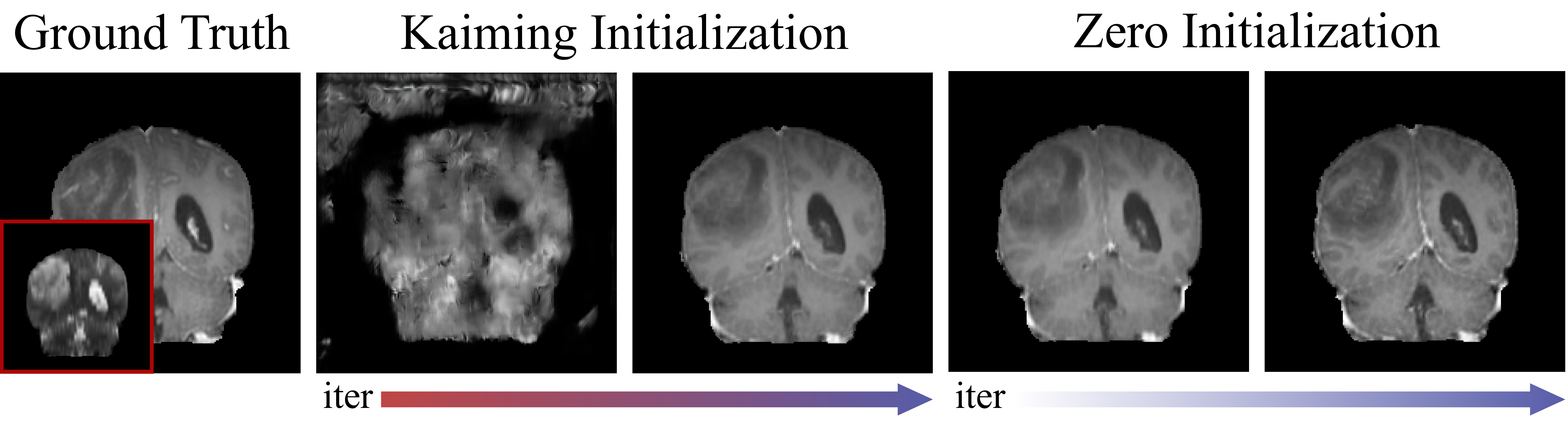}
    \vspace{3pt}
    \captionof{figure}{\textbf{Comparison of two initialization methods in the skip connection block for T2 $\rightarrow$ T1ce translation.} With Kaiming initialization, the image degrades rapidly early on. Zero initialization leads to a gradual increase in detail.}
    \label{fig9}
    \centering
    \captionof{table}{\textbf{Evaluation of generated report and VQA data}. GPT 4o is used to score the outputs on a scale from 0 to 10 based on predefined criteria.}
    \label{table3}
    \setlength{\tabcolsep}{12pt} 
    \centering
    \scalebox{0.75}{
        \begin{tabular}{l|cc}
            \toprule
            Model & Report generation & VQA generation \\
            \cmidrule(lr){1-3}
            Gemini 1.5 8B & 6.275 \small $\pm$ 2.667 & 8.008 \small $\pm$ 2.351 \\
            LLaMA 3.2 11B & 5.189 \small $\pm$ 1.799 & 6.837 \small $\pm$ 1.848 \\
            GPT 4o-mini  & 7.232 \small $\pm$ 2.106 & 8.025 \small $\pm$ 1.284 \\
            \midrule
            \textbf{Refined Data}  & \textbf{7.939 \small $\pm$ 2.174} & \textbf{8.538 \small $\pm$ 0.875} \\
            \bottomrule
        \end{tabular}
    }
    \centering
    \vspace{-6pt}
\end{table}

\vspace{3pt}
\noindent \textbf{Text Data Generation with LLMs.} We evaluated text data generated by our instruction-driven pipeline, which was designed to produce text data for image-only datasets. In this evaluation, we provided our generated text and evaluation criteria (from BioMed-VITAL \cite{cui2024biomedical}) to GPT 4o \cite{achiam2023gpt}, which scored the outputs on a scale from 0 to 10. Table \ref{table3} shows the average scores for generated reports and VQA outputs across different models. Among the individual language models, GPT 4o-mini achieved the highest score. When outputs from all three models were combined, the average score improved. 

\vspace{3pt}
\noindent \textbf{Ablation Study.} We performed an ablation study on each component of our method for image translation tasks with the results shown in Table \ref{table4}. Although using only the autoregressively generated image tokens from Stage I is sufficient for image-to-image synthesis, adding a learnable convolutional block significantly improved performance. In addition, zero initialization allowed the model to learn gradually, and both BiomedCLIP \cite{zhang2023biomedclip} and prompt tuning helped improve performance by allowing the model to adapt its features according to the target.

\begin{table} [t]
    \vspace{-6pt}
    \caption{\textbf{Ablation study results on the UPENN-GBM}. }
    \label{table4}
    \vspace{3pt}
    \setlength{\tabcolsep}{8pt} 
    \centering
    \scalebox{0.725}{
        \begin{tabular}{l|cccc|cccc}
            \toprule
            & \multicolumn{4}{c}{Segmentation} & \multicolumn{4}{|c}{T1 $\rightarrow$ T2} \\
              & WT. & ET. & TC. & Avg. & PSNR $\uparrow$ & SSIM $\uparrow$ & FID $\downarrow$ & LPIPS $\downarrow$  \\
            \cmidrule(lr){1-9}
            Stage I only      & 0.645 & 0.439 & 0.318 & 0.467 & 21.023 & 0.707 & 45.382 & 0.146 \\
            + Skip Conv block & 0.705 & 0.617 & 0.515 & 0.612 & 21.811 & 0.757 & 48.018 & 0.138 \\
            + Zero init       & 0.725 & 0.653 & 0.534 & 0.637 & 22.312 & 0.778 & 45.723 & 0.124 \\
            + Prompt tuning   & 0.773 & 0.762 & 0.608 & 0.714 & 22.617 & 0.801 & 45.047 & 0.117 \\
            + BioMedCLIP      & \textbf{0.792} & \textbf{0.778} & \textbf{0.613} & \textbf{0.727} & \textbf{22.943} & \textbf{0.817} & \textbf{43.205} & \textbf{0.111} \\
            \bottomrule
        \end{tabular}
    }
    \vspace{-6pt}
\end{table}

% \noindent \textbf{Nucleus Sampling.} 

\vspace{3pt}
\noindent \textbf{Impact of Zero Initialization.} We compared how different parameter initializations affect the training of the skip connection block. Fig. \ref{fig9} shows the comparison between the Kaiming initialization \cite{he2015delving} and the zero initialization. When Stage II starts with non-zero initial values, these early values tend to dominate, whereas zero initialization allows the model to gradually update its parameters while preserving the image features learned from the Stage I tokens. This approach not only improves performance but leads to more stable training.

\section{Discussion}

We have unified diverse medical tasks, from text generation to image translation and segmentation, within a single LLM framework. Our work shows that a single model can handle both image-to-text and image-to-image tasks while preserving the LLM's core linguistic reasoning. Although our use of synthetically generated text lacks clinical validation, we confirmed the model's fundamental understanding of medical imaging concepts. This limitation can be addressed through future training on clinician-annotated data. Furthermore, while our experiments were confined to brain MRI, we argue that our method is inherently extensible. Its strong performance suggests potential for application across other anatomical regions and modalities, provided it undergoes appropriate fine-tuning.

\vspace{3pt}
\noindent {\small \textbf{Acknowledgement.} This study was supported by the SMC-SKKU Future Convergence Research Program Grant (SMO125123); and the AI Graduate School Support Program (Sungkyunkwan University) (RS-2019-II190421), the National Research Foundation of Korea (RS-2024-00408040), the ICT Creative Consilience Program Grant (IITP-2026-RS-2020-II201821) and the AI Computing Infrastructure Enhancement (GPU Rental Support) User Support Program (RQT-25-090077), all of which were funded by the Ministry of Science and ICT (MSIT), Republic of Korea.}

% Acknowledgments
% \section{Acknowledgments}
% This study was supported by the National Research Foundation of Korea (RS-2024-00408040); the AI Graduate School Support Program (Sungkyunkwan University) (RS-2019-II190421); the ICT Creative Consilience program (IITP-2025-RS-2020-II201821); and the Artificial Intelligence Innovation Hub program (RS-2021-II212068).

% Reference 
{
    \small
    \bibliographystyle{splncs04}
    \bibliography{ref}
}

\clearpage
\appendix

\setlength{\abovedisplayskip}{\baselineskip}
\setlength{\belowdisplayskip}{\baselineskip}
\renewcommand{\thesection}{\Alph{section}}
\setcounter{section}{0}
\setcounter{footnote}{0}

\onecolumn

\section{Appendix}

\section{Dataset}
\label{sec:A}

\noindent \textbf{Preprocessing.} All 3D images were resampled to a uniform resolution of $1\times1\times1 \text{mm}^3$. Then, 2D slices centered on the region of interest (ROI) were extracted from the axial, sagittal, and coronal views. Each slice was cropped to $192\times192$ pixels and if an image was smaller, zero padding was applied to reach the required size.  From the center of the ROI, 3 to 5 2D slices were extracted.

\vspace{3pt}
\noindent \textbf{BraTS2021} \cite{baid2021rsna} includes multimodal MRI data from 1251 glioma patients. It  has four imaging modalities: T1-weighted, T1 contrast-enhanced, T2-weighted, and FLAIR images, with image dimensions of $240\times240\times155$ and a voxel spacing of $1\times1\times1 \text{mm}^3$. The ROIs included edema, enhancing tumor, and tumor core. Of the 1251 subjects, 1141 were used for training and 110 for evaluation.

\vspace{3pt}
\noindent \textbf{BraTS2023-MEN} \cite{labella2023asnrmiccai} contains multimodal MRI scans from 1000 meningioma patients. It includes T1-weighted, T1 contrast-enhanced, T2-weighted, and FLAIR images, with dimensions of $240\times240\times155$ and a voxel spacing of $1\times1\times1 \text{mm}^3$. The ROIs include hyperintensity, enhancing tumor, and tumor core. Out of the 1000 subjects, 909 were used for training and 91 for evaluation.

\vspace{3pt}
\noindent \textbf{IXI} \footnote{https://brain-development.org/ixi-dataset/} consists of normal brain MR images. A total of 578 subjects were used in this study. It includes three modalities: T1-weighted, T2-weighted, and Proton Density, with image dimensions of $256\times256\times150$ and a voxel spacing of $0.937\times0.9375\times1.2 \text{mm}^3$. Out of the 578 subjects, 527 were used for training and 51 for evaluation.

\vspace{3pt}
\noindent \textbf{ATLAS 2.0} \cite{liew2022large} contains brain MRI data from stroke patients, including T1-weighted images and manually segmented lesion masks from 496 subjects. The T1-weighted MRI data were acquired using 1.5 Tesla and 3 Tesla MR scanners and all images have a voxel spacing of $1\times1\times1 \text{mm}^3$.

\vspace{3pt}
\noindent \textbf{UPENN-GBM} \cite{bakas2021multi} includes multimodal MR images and corresponding ROIs. A total of 147 glioma subjects were used in this study. It includes T1-weighted, T1 contrast-enhanced, T2-weighted, and FLAIR images with dimensions of $240\times240\times155$ and a voxel spacing of $1\times1\times1 \text{mm}^3$. The ROIs encompass edema, enhancing tumor, and tumor core regions.

\section{Implementation Details.}
\label{sec:B}

\noindent \textbf{Base Language Model.} Our base language model is \texttt{dolly-v2-3b} \cite{dolly2023introducing}, which originally contained 50,821 token types ($K_\texttt{text}$). We have expanded its token embedding table to 51,845 entries by adding 1,024 new image tokens ($K_\texttt{img}$). For each image token generated by the VQ-GAN \cite{esser2021taming} encoder, we add $K_\texttt{text}$ to its value and pass it to the language model as a token ID. If the model outputs a token with an ID greater than or equal to $K_\texttt{text}$, it is treated as an image token, in which case we subtract $K_\texttt{text}$ from the token ID before sending it to the VQ-GAN decoder. The model was trained using the AdamW \cite{loshchilov2018decoupled} optimizer with a batch size of 2 and a learning rate of $5\times10^{-6}$

\vspace{3pt}
\noindent \textbf{VQ-GAN.} We used VQ-GAN as our image tokenizer and trained it on $192\times192$ images. The codebook contains 1024 indices and each embedding has 256 dimensions. The encoder and quantizer converted a $192\times192$ image into a $16\times16$ matrix, which then flattened the results in 144 image tokens. The model was trained using the Adam \cite{KingmaB14} optimizer with a batch size of 2 and a learning rate of $4.5\times10^{-6}$

\vspace{3pt}
\noindent \textbf{Skip connection.} We added a skip connection block to the existing VQ-GAN and kept the VQ-GAN frozen during training. The skip connection block used inputs from all feature maps at every scale and was made up of group normalization, SiLU \cite{elfwing2018sigmoid}, a convolution layer, and cross-attention. These skip connections were added to the decoder’s feature maps, and for segmentation, the decoder’s output was passed through an extra convolution layer before producing the final result. The model was trained using the Adam optimizer with a batch size of 2 and a learning rate of $4.5\times10^{-6}$

\section{Data Generation}
\label{sec:C}
This section describes the prompt and LLM used for data generation. The prompt was based on Biomed-VITAL \cite{cui2024biomedical}.

\vspace{3pt}
\noindent \textbf{Simple Caption.} We use the LLMs to generate text data by summarizing each image with a simple caption. An example of an image caption used is shown in Fig. \ref{figA}. During data generation, the LLM is provided with an instruction, an image, and its caption. Afterward, it generates a response accordingly.

\vspace{3pt}
\noindent \textbf{Prompt for Text Data Generation.} Similar to the prompt used in Biomed-VITAL \cite{cui2024biomedical}, a detailed prompt for generating report data is shown in Fig. \ref{figB}. For generating VQA, data are shown in Fig. \ref{figC}. This prompt is fed into Gemini 1.5 Flash-8B \cite{team2023gemini}, LLaMA 3.2 11B \cite{grattafiori2024llama3herdmodels}, and GPT 4o-mini \cite{achiam2023gpt}, and the final refined report is generated by GPT 4o-mini.

\vspace{3pt}
\noindent \textbf{Evaluation Prompt.} We rate the text data produced by the LLM on a scale from 0 to 10 using a prompt that includes multiple evaluation criteria along with GPT 4o. The evaluation prompt is shown in Fig. \ref{figD}.

\section{Instruction Template}
\label{sec:D}

\noindent The instructions used for \textbf{report generation} are illustrated in Fig. \ref{figE}.

\vspace{3pt}
\noindent The instructions used for \textbf{segmentation} are illustrated in Fig. \ref{figF}.

\vspace{3pt}
\noindent The instructions used for \textbf{image translation} are illustrated in Fig. \ref{figG}.

\clearpage

\fvset{formatcom=\rmfamily\mdseries}
\begin{figure*}[t]
\centering
\begin{minipage}{1\linewidth}
    \begin{adjustbox}{scale=0.65}
    \begin{minipage}{1.52\columnwidth}
    \begin{Verbatim}[frame=single, numbersep=5pt,
                     commandchars=\\\{\}]
    - Captured using a \{\texttt{modality}\} scan, this image shows a \{\texttt{plane}\} section of the brain and demonstrates a \{\texttt{abnormality}\}.
    - This image was obtained using a \{\texttt{modality}\} scan and presents a \{\texttt{plane}\} view of the brain. It illustrates a \{\texttt{abnormality}\}.
    - The \{\texttt{plane}\} view in this image is captured via a \{\texttt{modality}\} scan of the brain. It reveals a \{\texttt{abnormality}\}.
    - Utilizing a \{\texttt{modality}\} scan, this brain image displays a \{\texttt{plane}\} perspective. The image indicates a case of \{\texttt{abnomrality}\}.
    - This brain scan employs a \{\texttt{modality}\} modality to produce a \{\texttt{plane}\} view, highlighting a \{\texttt{abnormality}\}.
    \end{Verbatim}
    \end{minipage}
    \end{adjustbox}
    \centering
    \vspace{-6pt}
    \caption{The formats of simple caption for LLM to generate text data.}
    \label{figA}
\end{minipage}

\vspace{12pt}

\centering
\begin{minipage}{1\linewidth}
    \centering
    \begin{adjustbox}{scale=0.825}
    \begin{minipage}{1.2\columnwidth}
    \begin{Verbatim}[frame=single, numbersep=5pt,
                     commandchars=\\\{\}, breaklines=true,  
                     breaksymbolleft={}, breaksymbolright={}]
\textsf{\textcolor{blue}{messages}} = [\{\textsf{"type": "text", "text":} 
    "You are an AI radiologist specialized in biomedical topics. You are provided with a brain MR image. In some cases, you may have additional text (image context) that mentions the image. Please meticulously extract all possible visual details from the image, and when generating questions and answers, ensure to integrate and consider the provided supplementary textual information. It is crucial to highlight the connections and correlations between the textual content and the visual elements within the picture to capture the full context.

    Your task is to generate a report about the image. It is essential to thoroughly consider and reference the accompanying textual information (image caption and image context) to ensure that the answers highlight the significance of the visual details present.

    Below are the requirements for generating the questions and answers in the conversation:
    - Avoid quoting or referring to specific facts, terms, abbreviations, dates, numbers, or names, as these may reveal the conversation is based on the text information, rather than the image itself. Focus on the visual aspects of the image that can be inferred without the text information.
    - Ensure that questions are diverse and cover a range of visual aspects of the image.
    - Answer responsibly, avoiding overconfidence, and do not provide medical advice.
    - The report should be generated without any sub-items or breakdowns and should not contain any line breaks. It should also avoid using special characters, except for parentheses, commas, question marks, periods, exclamation points, etc. and the language should be English only.  
    - Extract as much key detailed information from the image as possible, and I will also provide you with some text to supplement the image.
    - Avoid hallucination and ensure that the generated report is relevant to the image and the text provided."\}]

\textsf{for \textcolor{blue}{sample} in \textcolor{blue}{fewshot\_samples}}:
    \textsf{\textcolor{blue}{messages}.append({"type": "image", "image": \textcolor{blue}{sample['image']}})}
    \textsf{\textcolor{blue}{messages}.append({"type": "text", "text": \textcolor{blue}{sample['report']}})}

\textsf{\textcolor{blue}{messages}.append({{"type": "image", "image": \textcolor{blue}{image}}})}
\textsf{\textcolor{blue}{messages}.append({{"type": "text", "text": \textcolor{blue}{caption}}})}
    \end{Verbatim}
    \end{minipage}
    \end{adjustbox}
    \centering
    \vspace{-6pt}
    \caption{The prompt for report data generation.}
    \label{figB}
\end{minipage}
\end{figure*}

\begin{figure*}[t]
\centering
\begin{minipage}{1\linewidth}
    \begin{adjustbox}{scale=0.825}
    \begin{minipage}{1.2\columnwidth}
    \begin{Verbatim}[frame=single, numbersep=5pt,
                     commandchars=\\\{\}, breaklines=true,  
                     breaksymbolleft={}, breaksymbolright={}]
\textsf{\textcolor{blue}{messages}} = [\{\textsf{"type": "text", "text":} 
    "You are an AI radiologist specialized in biomedical topics.
    You are provided with a brain MR image. In some cases, you may have additional text (image Context) that mentions the image. Please meticulously extract all possible visual details from the image, and when generating questions and answers, ensure to integrate and consider the provided supplementary textual information. It is crucial to highlight the connections and correlations between the textual content and the visual elements within the picture to capture the full context.

    Your task is to generate a conversation between a person (User) inquiring about the image and you (Assistant) responding to their questions. During this interaction, the conversation should evolve as if both the User and Assistant are observing the image together. It is essential to thoroughly consider and reference the accompanying textual information (image caption and image context) to ensure a rich and informative exchange that highlights the significance of the visual details present.

    Below are the requirements for generating the questions and answers in the conversation:
    - Avoid quoting or referring to specific facts, terms, abbreviations, dates, numbers, or names, as these may reveal the conversation is based on the text information, rather than the image itself. Focus on the visual aspects of the image that can be inferred without the text information.
    - Ensure that questions are diverse and cover a range of visual aspects of the image.
    - The conversation should encompass a minimum of 3-4 exchanges of questions and answers. You may adjust the number of rounds based on the provided image and text. For content with substantial information, employing additional questions and answers may be more appropriate to ensure thorough discussion and understanding.
    - Answer responsibly, avoiding overconfidence, and do not provide medical advice. 
    - You need to start by generating the User's questions first, not assistant's answer.
    - Extract as much key detailed information from the image as possible, and I will also provide you with some text to supplement the image.
    - Avoid hallucination and ensure that the generated report is relevant to the image and the text provided.
    
    You can use the following example Questions as a guide to generate the conversation:
    - What type of scan is used for this image?
    - What abnormality do you see in this image?
    - What is the orientation of this image?"\}]

\textsf{for \textcolor{blue}{sample} in \textcolor{blue}{fewshot\_samples}}:
    \textsf{\textcolor{blue}{messages}.append({"type": "image", "image": \textcolor{blue}{sample['image']}})}
    \textsf{\textcolor{blue}{messages}.append({"type": "text", "text": \textcolor{blue}{sample['conversations']}})}

\textsf{\textcolor{blue}{messages}.append({{"type": "image", "image": \textcolor{blue}{image}}})}
\textsf{\textcolor{blue}{messages}.append({{"type": "text", "text": \textcolor{blue}{caption}}})}
    \end{Verbatim}
    \end{minipage}
    \end{adjustbox}
    \centering
    \vspace{-6pt}
    \caption{The prompt for VQA data generation.}
    \label{figC}
\end{minipage}

\end{figure*}

\begin{figure*}[t]
\vspace{-4pt}
\centering
\begin{adjustbox}{scale=0.825}
\begin{minipage}{1.2\columnwidth}
\begin{Verbatim}[frame=single, numbersep=5pt,
                 commandchars=\\\{\}, breaklines=true,  
                 breaksymbolleft={}, breaksymbolright={}]
\textcolor{blue}{\textsf{messages}} = [\{\textsf{"type": "text", "text": }
    Assume that you are a medical expert with extensive experience in your field. Your task is to assess and score a set of question-and-answer pairs from an instruction following dataset designed for fine-tuning a medical large language model (LLM). You should give a score for each Q-A pair as you are provided with multi AI conversations. You will be provided with images, their corresponding captions, in-text mentions, and the Q&A pairs. Your scoring, ranging from 0 to 10, will evaluate the following criteria:
    - Scope and Relevance: How well does the question cover key aspects of the medical image and caption provided?
    - Value for Fine-Tuning: Is the question formulated in a way that it will add value to the fine-tuning process of the medical LLM?
    - Answer Alignment: Does the provided answer directly address the question posed?
    - Accuracy: Is the information in the answer medically accurate and correct?
    - Utility: How useful is the answer in a medical context? Does it provide actionable or insightful information?
    - Image Content Recognition and Utilization: Does the response accurately identify the content depicted in the image and effectively incorporate this information into the answer to enhance comprehension or applicability in a medical context?
    
    Please consider additional factors such as:
    - Clarity: Are both the question and the answer clearly articulated and free of ambiguity?
    - Detail and Depth: Do the answer's details contribute to a deeper understanding of the topic?
    - Medical Precision: How precisely do the question and answer reflect medical terminology and knowledge?
    
    As you review each Q&A pair, please first output a single line containing scores of each Q&A pairs, splited by blank. The first score is for the first question and its corresponding answer, and so on. After that you can give some explanations, like what are the shortcomings of the current instructions.\},
    \{\textsf{"type": "image", "image": \textcolor{blue}{image}}\},
    \{\textsf{"type": "text", "text": \textcolor{blue}{caption}\}}]
    
\textsf{for  \textcolor{blue}{vqa\_data} in  \textcolor{blue}{vqa\_datas}:}
    \textsf{ \textcolor{blue}{messages}.append}(\{\textsf{"type": "text", "text":} \textcolor{blue}{\textsf{vqa\_data}}\})
\end{Verbatim}
\end{minipage}
\end{adjustbox}
\centering
\vspace{-7pt}
\caption{The prompt for data evaluation.}
\label{figD}
\vspace{-3pt}
\end{figure*}

\begin{figure*}[t]
\centering
\begin{adjustbox}{scale=0.825}
\begin{minipage}{1.2\columnwidth}
\begin{Verbatim}[frame=single, numbersep=5pt,
                 commandchars=\\\{\}]
    - Generate free-text radiology reports for the provided brain MR images.
    - Use the provided brain MR images to create corresponding free-text radiology reports.
    - Based on the provided brain MR images, produce free-text radiology reports.
    - Create free-text radiology reports that correspond to the provided brain MR images.
    - Utilize the provided brain MR images to generate corresponding free-text radiology reports.
    - Generate free-text radiology reports in accordance with the provided brain MR images.
    - Use the provided brain MR images to create accurate free-text radiology reports.
    - Produce free-text radiology reports that match the provided brain MR images.
    - Create free-text radiology reports that are consistent with the provided brain MR images.
    - Utilize the provided brain MR images to generate comprehensive free-text radiology reports.
\end{Verbatim}
\end{minipage}
\end{adjustbox}
\centering
\vspace{-7pt}
\caption{Instruction list for report generation task.}
\label{figE}
\vspace{-3pt}
\end{figure*}

\begin{figure*}[t]
\centering
\begin{adjustbox}{scale=0.825}
\begin{minipage}{1.2\columnwidth}
\begin{Verbatim}[frame=single, numbersep=5pt,
                 commandchars=\\\{\}]
    - Generate a segmentation map for \{\texttt{roi}\} in a given brain MR image.
    - Create a segmentation of \{\texttt{roi}\} in the brain MR image that highlights key structures.
    - Produce a segmented brain MR image identifying \{\texttt{roi}\}-related regions and structures.
    - Use the provided brain MR image to create the corresponding \{\texttt{roi}\} segmentation map.
    - Generate a brain MR image segmentation that accurately delineates regions of \{\texttt{roi}\}.
    - Based on the input brain MR image, produce a detailed \{\texttt{roi}\} segmentation map.
    - Use the input brain MR image to generate a segmentation highlighting areas of \{\texttt{roi}\}.
    - Create a segmentation of \{\texttt{roi}\} in the brain MR image corresponding to the input image.
    - Produce a brain MR image segmentation map that identifies regions of \{\texttt{roi}\}.
    - Generate a \{\texttt{roi}\} segmentation map from the given brain MR image to enhance visualization.
\end{Verbatim}
\end{minipage}
\end{adjustbox}
\centering
\vspace{-7pt}
\caption{Instruction list for segmentation task.}
\label{figF}
\vspace{-3pt}
\end{figure*}

\begin{figure*}[t]
\centering
\begin{adjustbox}{scale=0.825}
\begin{minipage}{1.2\columnwidth}
\begin{Verbatim}[frame=single, numbersep=5pt,
                 commandchars=\\\{\}]
    - Transform the brain MR image from \{\texttt{source}\} to \{\texttt{target}\}, 
      ensuring the output reflects the characteristics of \{\texttt{target}\}.
    - Convert the input brain MR image from \{\texttt{source}\} into a brain MR image in \{\texttt{target}\}.
    - Use the brain MR image in \{\texttt{source}\} to generate its corresponding image in \{\texttt{target}\}.
    - Generate a brain MR image in \{\texttt{target}\} based on the input from \{\texttt{source}\}.
    - Create a \{\texttt{target}\} brain MR image that mirrors the characteristics of the \{\texttt{source}\}.
    - Produce a brain MR image in \{\texttt{target}\} using the provided \{\texttt{source}\} image.
    - Transform the input brain MR image from \{\texttt{source}\} into the corresponding image in \{\texttt{target}\}.
    - Based on the provided brain MR image in \{\texttt{source}\}, create the equivalent image in \{\texttt{target}\}.
    - Generate a brain MR image in \{\texttt{target}\} that aligns with the features of the \{\texttt{source}\} image.
    - Synthesize a brain MR image in \{\texttt{target}\} using the input image from \{\texttt{source}\}.
\end{Verbatim}
\end{minipage}
\end{adjustbox}
\centering
\vspace{-7pt}
\caption{Instruction list for image translation task.}
\label{figG}
\vspace{-4pt}
\end{figure*}

\end{document}